\documentclass[conference,10pt,times,mathptm,psfig]{IEEEtran}
\usepackage{times}
\usepackage{epsfig}
\usepackage{graphicx}
\usepackage{amsmath}
\usepackage{amssymb}
\usepackage{pifont}
\usepackage{bm}
\usepackage{balance}
\usepackage{booktabs}
\usepackage{multirow}
\usepackage{threeparttable}
\usepackage{bbding}
\usepackage{hyperref}
\usepackage{cite}

\begin{document}
%
\title{Part-level Action Parsing via a Pose-guided Coarse-to-Fine Framework}


\author{Xiaodong Chen$^1$\quad
Xinchen Liu$^2$ \quad
Wu Liu$^2$ \quad
Kun Liu$^2$ \quad
Dong Wu$^2$ \quad
Yongdong Zhang$^1$ \quad
Tao Mei$^2$ \\
$^1$University of Science and Technology of China, Hefei, China \quad $^2$JD.com, Beijing, China \\
{\tt\footnotesize cxd1230@mail.ustc.edu.cn,\{liuxinchen1,liuwu1,liukun167,wudong\}@jd.com,zyd73@ustc.edu.cn,tmei@live.com}
}

\maketitle

\begin{abstract}
Action recognition from videos, i.e., classifying a video into one of the pre-defined action types, has been a popular topic in the communities of artificial intelligence, multimedia, and signal processing.
However, existing methods usually consider an input video as a whole and learn models, e.g., Convolutional Neural Networks (CNNs), with coarse video-level class labels.
These methods can only output an action class for the video, but cannot provide fine-grained and explainable cues to answer why the video shows a specific action.
Therefore, researchers start to focus on a new task, Part-level Action Parsing (PAP), which aims to not only predict the video-level action but also recognize the frame-level fine-grained actions or interactions of body parts for each person in the video.
To this end, we propose a coarse-to-fine framework for this challenging task.
In particular, our framework first predicts the video-level class of the input video, then localizes the body parts and predicts the part-level action.
Moreover, to balance the accuracy and computation in part-level action parsing, we propose to recognize the part-level actions by segment-level features.
Furthermore, to overcome the ambiguity of body parts, we propose a pose-guided positional embedding method to accurately localize body parts.
Through comprehensive experiments on a large-scale dataset, i.e., Kinetics-TPS, our framework achieves state-of-the-art performance and outperforms existing methods over 31.10\% ROC score.

\end{abstract}

\begin{IEEEkeywords}
Part-level Action Parsing, Action Recognition, Video Understanding, Pose-Guided Positional Embedding
\end{IEEEkeywords}

%
\IEEEpeerreviewmaketitle

\section{Introduction}

Action recognition~\cite{journals/corr/KayCSZHVVGBNSZ17, conf/iccv/GoyalKMMWKHFYMH17, conf/cvpr/GuSRVPLVTRSSM18, conf/iscas/HouZZS14, conf/iscas/ZhuHFCSL21,journals/tmm/LuWMGF14}, which can be treated as a high-level video classification problem, is a hot topic of video understanding in computer vision. 
With the development of rich representations based on neural networks~\cite{conf/cvpr/HeZRS16, journals/cacm/KrizhevskySH17, conf/iccv/QiuYM17}, significant progress has been made on this task. 
Although these existing action recognition methods~\cite{conf/eccv/WangXW0LTG16, conf/iccv/LinGH19, conf/iccv/Feichtenhofer0M19, journals/tmm/LiYDMR19} can predict the high-level human action of the whole video, they neglect the detailed and middle-level understanding of human actions. 

\begin{figure}
\begin{center}
\includegraphics[width=0.95\linewidth]{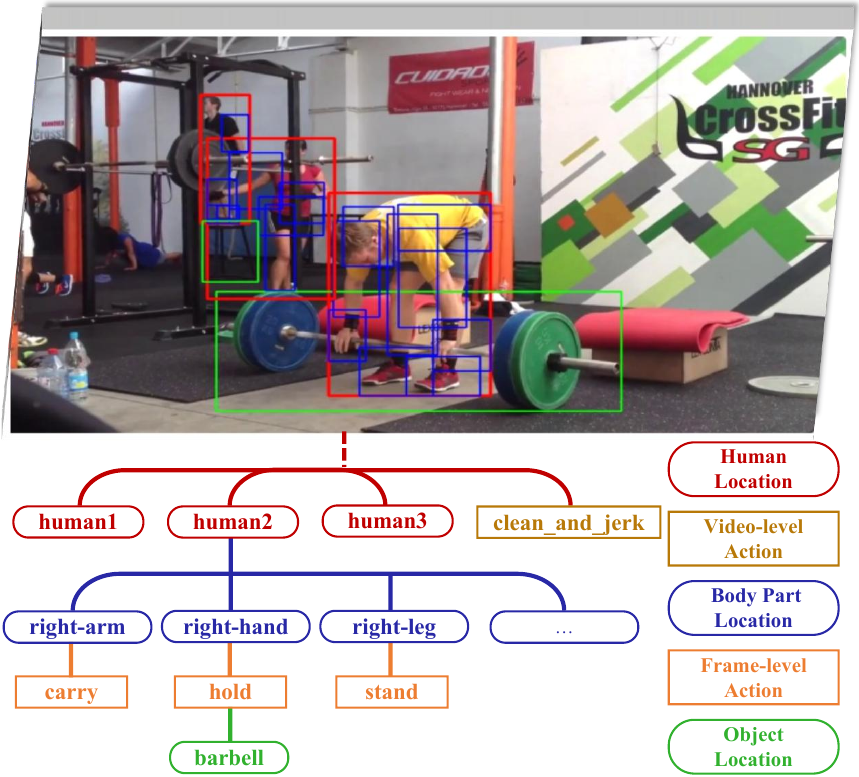}
\end{center}
   \caption{An example of the Part-level Action Parsing (PAP) task. The upper part is the input video, and the lower part is the labels need to predict, which include  video-level action, location of each person, location of each body part, and frame-level action of each body part.}
\label{fig:figure1} \vspace{-7mm}
\end{figure}

To fill this gap, the Part-level Action Parsing (PAP) task, which aims to recognize frame-level human action of all body parts and the whole body from a video in the wild, was firstly focused on by researchers recently. 
The PAP task is to address the following problem: given a human action video, a system needs to predict the human location, body part location, part state/action in each frame, then integrates these results to predict human action in the video level.
For instance, as video of the fitness center shown in Fig.~\ref{fig:figure1}, we not only need to predict its video-level label as ``clean\_and\_jerk'', but also need to detect each body part such as ``right\_arm'' and ``right\_hand'' in each frame and predict their part-level action labels like ``carry'' and ``hold''.

By decomposing an action into a human part graph, the PAP task advances the area of human action understanding with a shift from the traditional action recognition task to deeper understanding tasks of part-level action parsing.
Moreover, it may have many potential applications such as intelligent manufacturing, sports analysis, fitness instruction, and so on.
Therefore, this paper concentrates on the part-level action parsing task which is valuable yet overlooked by the community of multimedia and computer vision.

However, part-level action parsing is a non-trivial task that faces several challenges as shown in Fig.~\ref{fig:figure1}. 
First of all, there are many obstacles when predicting the spatial position of the human body and body parts accurately.
Different from traditional object detection tasks, body part detection needs to overcome the ambiguity of body parts and capture the prior of human body structure.
Moreover, even if there are only two or three people in every frame of the video, the number of part-level actions that need to be predicted is very large due to the fine-grained division of human body parts. 
It is more strenuous to represent the relationship between these parts.
Furthermore, the trade-off between computational power and accuracy in prediction should also be considered due to the densely predicted frame-level and part-level actions.

\begin{figure*}
\begin{center}
\includegraphics[width=0.95\linewidth]{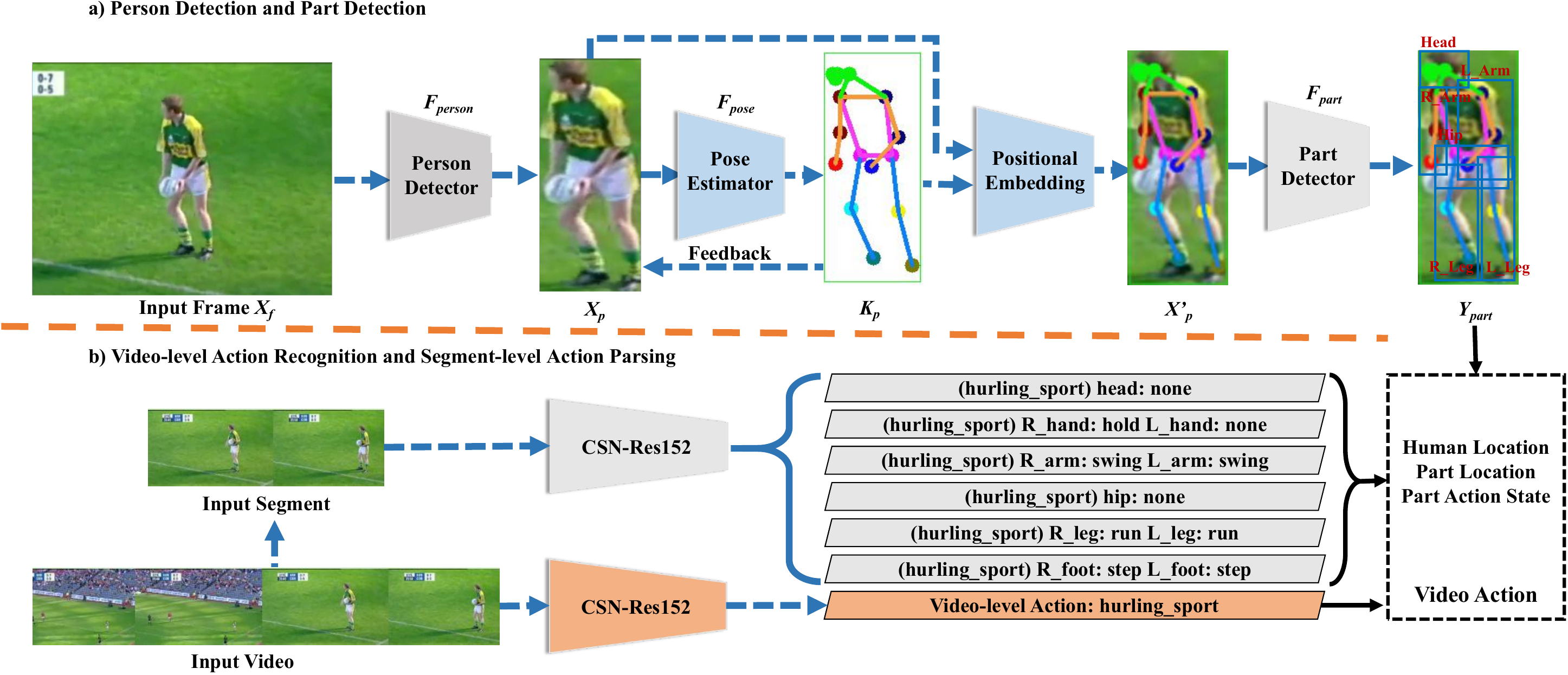}
\end{center}
   \caption{The overall architecture of our method, which includes two main parts: a) person detection and part detection, and b) video-level action recognition and segment-level action parsing.}
\label{fig:figure2}
\vspace{-5mm}
\end{figure*}

Action recognition has been studied for several years. 
From data-driven representations learned by deep Convolutional Neural Network (CNN)~\cite{conf/iccv/QiuYM17, conf/iccv/TranBFTP15, conf/iscas/KimAS19, conf/aaai/LiuLGTM18} to Transform-based Neural Network~\cite{conf/iccv/LiuLCHW21} with large parameters, the accuracy of action recognition has been significantly improved. 
However, traditional action recognition methods often consider the whole video or clip as the smallest unit.
Despite their excellent performance for video-level action recognition, these methods cannot work well for frame-level part actions.
In addition, although some researchers have studied frame-level human action location~\cite{conf/cvpr/GuSRVPLVTRSSM18}, they only focus on the whole body action and ignore the fine-grained part-level actions. Due to the small size of body parts, the traditional methods, e.g., RoIAlign~\cite{conf/iccv/HeGDG17}, that used in frame-level human action location have little performance improvement on the PAP task.

To this end, we propose a Pose-guided Coarse-to-Fine framework, named PCF, for part-level action parsing. 
We first adopt the existing action recognition methods, e.g., CSN~\cite{conf/iccv/TranWFT19}, to predict the coarse action of the whole video, since it is the State-of-The-Art (SoTA) CNN-based model in the action recognition task. After that, we predict the fine-grained segment-level body part action instead of the frame-level action based on the persistence of human actions, which greatly improves the computational efficiency with less precision reduction. 
Moreover, due to the ambiguity of body parts, e.g., the similarity of the appearance of the left leg and the right leg, traditional existing object detectors are often unable to predict the body part effectively. 
To solve this problem, we propose the pose-guided positional embedding method which guides the detector to predict the part locations with human pose keypoints. 
By encoding each human keypoints with different colored dots on the original images, the feature representations of different parts are more easily distinguished by the detector, which effectively reduces the body part ambiguity.

In summary, the contributions of this paper include: 
1) we make one of the first attempts for part-level action parsing which is a valuable yet unexplored task; 
2) we design a PCF framework to exploit the potential performance of existing object detectors with pose-guided positional embedding and predict both the coarse video-level action and the fine-grained body part action; 
3) our method achieves SoTA results on the Kinetics-TPS dataset\cite{github/KineticsTPS/Hypnosx21}, which shows the effectiveness of our method.

\section{THE PROPOSED FRAMEWORK}
\label{sec:method}

\subsection{Overall Framework}
\label{subsec:framework}

Figure~\ref{fig:figure2} shows the overall structure of the Pose-guided Coarse-to-Fine (PCF) framework for the PAP task.
It includes three stages, i.e., instance and part detection, video-level action recognition, and part action parsing.
In the first stage, as shown in the upper part of Figure~\ref{fig:figure2}, we adopt YOLOF~\cite{conf/cvpr/ChenWYZC021} as the backbone of the person detector and part detector to locate each person and body parts. 
To overcome the ambiguity of body parts, we insert the pose estimator and the positional embedding module between the person detector and part detector to improve the accuracy of the part location. 
The second stage is shown in the lower part of Figure~\ref{fig:figure2}. Based on the short-term persistence of human actions, we exploit segment-level action prediction to approximate the frame-level action state to balance the accuracy and computation cost.
In particular, we divide the original video into multiple segments that last three seconds or so.
Then tag each segment with six segment-level pseudo action labels based on the original frame-level part action labels which significantly reduce the computation cost for frame-level action parsing. 
After that, we train models for segment-level action and video-level action respectively.
In the final stage, we integrate the output of all previous stages to get the final output for the PAP task. Next, we will introduce each stage of our framework in detail.

\subsection{Pose-guided Part Detection}
\label{subsec:detection}
To our knowledge, body part detection is an unprecedented task in traditional object detection tasks.
Different from general object detection, body part detection needs to overcome the ambiguity of body parts and model the prior of human body structure. 
For example, normally people only have one left foot and one right foot, but the local features of the left one and the right one are often very similar.
Fortunately, this structural prior is very common in human pose estimation and has been widely explored. 
To maximize the usage of this structural prior, we propose a pose-guided part detection method, which is shown in the top half of Figure~\ref{fig:figure2}.

In detail, the person detector $F_{person}(\cdot)$ first extracts the bounding box of a person $X_p$ from the input frame $X_f$ by
\begin{equation}
X_{p} = F_{person}(X_{f}).
\label{equ:equ1}
\end{equation}
Then the pose estimator $F_{pose}(\cdot)$ takes $X_p$ as the input and outputs the keypoints $K_{p} = \{(x_i, y_i)\}^N_{i=1}$ of the person, which is formulated by 
\begin{equation}
K_{p} = F_{pose}(X_{p}),
\label{equ:equ2}
\end{equation}
where $(x_i, y_i)$ is the coordinates of keypoint $i$, $N$ is the number of keypoints.
After that, the keypoints $K_{p}$ are integrated by the positional embedding module $G$ with dots of different colors and radius on the original $X_p$ to generate an augmented person image $X'_p$.
This process can be formulated by
\begin{equation}
X'_{p} = X_{p} + G(K_{p}).
\label{equ:equ3}
\end{equation}
By this means, we can increase the appearance difference between different body parts and facilitate the learning of body parts detector $F_{body}$. 
Finally, the part detector $F_{body}$ is implemented to localize each body part $Y_{part}$ by
\begin{equation}
Y_{Part} = F_{body}(X'_{p}).
\label{equ:equ4}
\end{equation}

In addition, we also fine-tune the person detection box with the results of the pose estimator.
In a nutshell, the pose estimator has the ability to predict the possible human keypoints outside the person box, and we fine-tune the detected person box until all possible human keypoints are included.

\subsection{Part State Parsing and Action Recognition} 
\label{subsec:parsingtorecongnition}
Fine-grained frame-level part state parsing requires more computation and hardware cost than coarse video-level action recognition.
However, we find this frame-level action parsing problem can be transformed into a simpler segment-level action recognition task due to the overwhelming ``Long Tail Effect'' caused by the short-term persistence of actions in video segments.
For example, in the segment of ``hurling sport'' that lasts about three seconds, we just need to predict ``None'' for the heads in every frame and easily achieve 97.7\% frame-level part state accuracy. 
To take advantage of the significant ``Long Tail Effect'', as shown in the bottom half of Figure~\ref{fig:figure2}, we tag each video segment with six part-level pseudo labels based on the original frame-level action state label.
The fine-grained part-level label consists of three parts: coarse video-level action, body part, and the most frequent frame-level action of the body part. 
As the example shown in Figure~\ref{fig:figure2}, ``(hurling\_sport) head: none'' means the video-level action of this video is ``hurling\_sport'', and the most frequently frame-level action of the ``head'' in this segment is ``none''. 
Through this transformation, we can directly apply individual action recognition networks without sharing parameters, such as CSN~\cite{conf/iccv/TranWFT19}.
By this means, we can predict each fine-grained segment-level label and coarse video-level label respectively without any other models related to the computation-intensive frame-level action prediction.

\section{EXPERIMENTS}
\label{sec:experiments}

\subsection{Experimental Setting}
\label{subsec:details}

\begin{table*}[t]
\begin{center}
    \caption{Results of different settings on the Kinetics-TPS testing set. ``ROC Score'' refers to the final scores of the methods.} \vspace{-2mm}
    \label{tab:table1}
    \footnotesize
    \begin{threeparttable}
    \begin{tabular}{lcccc}
    Methods & Input & backbone & Video Acc (\%) & ROC Score (\%) \\
    \midrule
    baseline~\cite{github/deeperaction/kineticstps21} & RGB & TSN-Res50~\cite{conf/eccv/WangXW0LTG16} & - & 29.79 \\
    \midrule
    PCF (TSN\_RGB)  & RGB  & TSN-Res50 & 74.03  & 49.23 (+19.44) \\
    PCF (TSN\_Flow) & Flow & TSN-Res50 & 83.48  & 54.33 (+5.1)  \\
    PCF (Ours)       & RGB  & ip-CSN-152~\cite{conf/iccv/TranWFT19} & \textbf{96.46} & \textbf{60.89 (+6.56)}  \\
    \end{tabular}
    \end{threeparttable}
\end{center} \vspace{-7mm}
\end{table*}

\begin{table}[t]
\begin{center}
    \caption{Effect of Pose-guided Positional Embedding. }
    \vspace{-2mm}
    \label{tab:table2}
    \footnotesize
    \begin{threeparttable}
    \begin{tabular}{lccc}
    Model & Pose & $AP$ (\%) & $AP_{50}$ (\%) \\
    \midrule
    $YOLOF_{person}$ & \ding{55}    & 74.60 & 93.40 \\
    $YOLOF_{person}$ & \checkmark & 74.80 (\textbf{+0.20}) & 93.80 (\textbf{+0.40})  \\
    $YOLOF_{part}$   & \ding{55}    & 36.40 & 53.10 \\
    $YOLOF_{part}$   & \checkmark & 57.10 (\textbf{+20.7}) & 79.70 (\textbf{+26.6}) \\
    \end{tabular}
    \end{threeparttable} 
\end{center} \vspace{-5mm}
\end{table}

\textbf{Dataset.} 
The experiments are performed on the Kinetics-TPS dataset~\cite{github/KineticsTPS/Hypnosx21} that provides 7.9 M annotations of 10 body parts, 7.9 M part state (i.e., how a body part moves) of 74 body actions, and 0.5 M interactive objects of 75 objects in the video frames of 24 human action classes. 
Kinetics-TPS contains 3,809 training videos (4.96 GB in size) and 932 test videos (1.26 GB in size). 
It's worth noting the source videos of Kinetics-TPS come from Kinetics 700. Hence, all the Kinetics-pretrained models are forbidden in the PAP task.

\textbf{Evaluation Metrics.}
We adopt the official evaluation metric, i.e., ROC score, of the Kinetics-TPS dataset~\cite{github/KineticsTPS/Hypnosx21}.
ROC scores are calculated based on the Part State Correctness (PSC) and the action recognition conditioned on PSC.
The PSC calculates the accuracy of the whole human detection results and body part action parsing in each frame.
The action recognition conditioned on PSC draws the ROC curve and calculates the ROC score according to the top-1 video-level action recognition accuracy and PSC accuracy.
Please refer to~\cite{github/xiadingZ/evaluation} for more details.

\textbf{Details of Detector and Pose Estimator.} 
For person detector and part detector on keyframes, we adopted the YOLOF~\cite{conf/cvpr/ChenWYZC021}, which is an anchor-free model with a ResNet-101~\cite{conf/cvpr/HeZRS16} backbone. 
The model is pre-trained on the COCO dataset~\cite{conf/eccv/LinMBHPRDZ14} and then fine-tuned on Kinetics-TPS. 
The final models obtain 93.8 AP@50 in the person category and 79.7 AP@50 in the 10 body part categories on the Kinetics-TPS validation set.
For the pose estimator, we directly adopted the HRNet-w48~\cite{conf/cvpr/0009XLW19} pre-trained on COCO~\cite{conf/eccv/LinMBHPRDZ14} to extract the keypoints of each person without any fine-tuning.

\textbf{Details of Action Parsing and Action Recognition Network.} 
We use the CSN networks~\cite{conf/iccv/TranWFT19} as the backbone in our action recognition and action parsing framework.
We use the ip-CSN-152 implementation pre-trained on the IG-65M~\cite{conf/cvpr/GhadiyaramTM19} dataset with input sampling $T \times \tau = 32 \times 2$.
In particular, we freeze the Batch Normalization (BN) layers in the backbone during fine-tuning on Kinetics-TPS. 

\textbf{Details of Training.} 
The detection model and action recognition models are trained separately.
Each model is trained in an end-to-end manner.
In detail, we train the YOLOF detector using SGD with a mini-batch size of six on four V100 GPUs and train it for 24 epochs with a base learning rate of 0.01, which is decreased by a factor of 10 at epoch 16 and 22. 
We perform linear warm-up~\cite{journals/corr/GoyalDGNWKTJH17} during the first 1800 iterations.
For the CSN model, we train it using SGD with a mini-batch size of four on four V100 GPUs for 58 epochs with a base learning rate of 8e-5, which is decreased by a factor of 10 at epoch 32 and 48. 
We perform linear warm-up~\cite{journals/corr/GoyalDGNWKTJH17} during the first 16 iterations. 
By default, we use weight decay of 1e-4 and Nesterov momentum of 0.9 for all models.

\textbf{Details of Inference.} 
Following the official guideline~\cite{github/xiadingZ/evaluation}, we extract the top-10 results from the person detector and the top-1 results of each body part from the part detector during testing.
For the action recognition task and action state parsing task, we set the number of sampling clips as seven for each video segment at test time and scale the shorter side of input frames to 256 pixels.

\subsection{Main Results}
\label{subsec:results}
To demonstrate the effectiveness of our method, We compare our PCF framework with the official baseline method.
Meanwhile, to illustrate the fairness of comparison, we replace our ip-CSN-152 backbone with the TSN-Res50 used by the official baseline.

We present our results on Kinetics-TPS in Table~\ref{tab:table1}.
The "Input" in the second column refers to the video input form, and the calculation of optical flow is based on the tvl1 algorithm~\cite{journals/ipol/PerezMF13}.
The ``video acc'' in the fourth column refers to the top-1 video-level action recognition accuracy, while the ``ROC score'' in the fifth column refers to the final ROC score of the methods.

From the results, we can first find that directly applying our PCF framework with TSN-Res50 backbone and RGB input form, our performance achieves a significant enhancement of {+19.44} ROC score. 
Out of our expectation simply changing the input mode from RGB to optical flow gives a total boost of {+5.1} ROC score improvement. 
This may indicate that the body part action encoded by optical flow carries more effective information than RGB input when using 2D-CNN based network in the PAP task. 
Furthermore, with the strong CNN-based model ip-CSN-152 pretrained on IG-65M, our PCF framework achieves the {60.89}\% ROC score on the Kinetics-TPS dataset.

\begin{table}[t]
\begin{center}
    \caption{Frame-level Prediction v.s. Segment-level Prediction.}\vspace{-2mm}
    \label{tab:table3}
    \footnotesize
    \begin{threeparttable}
    \setlength{\tabcolsep}{1mm}{
    \begin{tabular}{l|c|cccccc|c}
    \multirow{2}{*}{Method}& \multirow{2}{*}{Duration} & \multicolumn{6}{c|}{The action prediction accuracy} & \multirow{2}{*}{TFLOPs} \\
    \cline{3-8} 
     & & head & arm & hand & hip & leg & foot  \\
    \midrule
    Frame & - & 92.46 & 68.11  & 68.50  & 84.19  & 66.27  & 65.47 & 36.12 \\
    Segment & 3.00s & 92.46 & 68.07  & 68.43  & 84.14  & 66.14  & 65.36 & 11.50\\
    Segment & 10.0s & 92.44 & 67.69  & 68.10  & 83.99 & 65.59  & 65.02 & 3.482 \\
    \end{tabular}}
    \end{threeparttable}
\end{center} \vspace{-5mm}
\end{table}

\subsection{Ablation Experiments}
\label{subsec:ablation}

\textbf{Effect of Adding Pose Estimator.} 
We investigate the effect of the pose estimator on detection mAP. 
For person detector and part detector, we train the lightweight CNN-based model YOLOF with human location and body parts location respectively. 
As shown in Table~\ref{tab:table2}, adding the pose estimator brings consistent AP and AP@50 increases for these two models. 
More specifically, equipped with the pose estimator, our $YOLOF_{part}$ model achieves a significant enhancement of $+26.6$ AP@50 on the Kinetics-TPS dataset.

\textbf{Frame-level Prediction v.s. Segment-level Prediction} 
In this subsection, we quantitatively compare the action prediction accuracy and the computation cost between the frame-level action parsing and the segment-level action parsing in Table~\ref{tab:table3}. 
The action prediction accuracy and the ``TFLOPs'' are calculated with the ip-CSN-152 model on the Kinetics-TPS dataset.
From the results, we can see that the computation ``TFLOPs'' decreases greatly with the increase of segment duration, while the loss of the accuracy is just 0.4\% or less. Especially when the duration of the segment is less than three seconds, the accuracy of frame-level prediction and segment-level prediction is almost the same (decrease less than 0.1\%), while the computation decreases about 68.16\%.

\section{CONCLUSION}
This paper presents a pose-guided coarse-to-fine framework for the part-level action parsing task. 
In our PCF framework, the pose-guided part detector is one of the first attempts toward body part detection and brings considerable improvement in the AP@50 (+26.60\%). Meanwhile, we convert the frame-level part state parsing problem into segment-level action recognition based on the persistence of human actions, which greatly improves the computational efficiency with less precision reduction. 
At last, our method achieves SoTA results at the Kinetics-TPS dataset, which shows the effectiveness of our PCF framework. With these three contributions, we provide one of the first attempts for the part-level action parsing task.

\section*{Acknowledgment}
This work was supported by the National Key R\&D Program of China under Grant No. 2020AAA0103800.

\newpage
{
\balance
\bibliographystyle{IEEEtran}
\bibliography{IEEEabrv}
}

\end{document}